%% file: paper2.tex
\documentclass[letterpaper, 10 pt, conference]{ieeeconf}  

\IEEEoverridecommandlockouts
\usepackage{amsmath}
\usepackage{amssymb}

\usepackage{cite}
\usepackage{algorithmic}
\usepackage{graphicx}
\usepackage{textcomp}
\usepackage[svgnames]{xcolor}
\usepackage{bm}
\usepackage{multirow}
\usepackage{siunitx}
\usepackage{tikz, pgfplots}
\usepackage{mathtools}
\usepackage{siunitx}
\usepackage{verbatim}
\usepackage{babel}
\usepackage{filecontents}
\usepackage{pgfplotstable}
\usepgfplotslibrary{colorbrewer}
\pgfplotsset{compat = 1.15, cycle list/Set1-8} 
\usetikzlibrary{pgfplots.statistics, pgfplots.colorbrewer} 
\usepackage{filecontents}
\usepackage{float} 
\usepackage{graphicx}
\usepackage{caption}
\usepackage{subcaption}
\usepackage{ifthen}
\usepackage{pgfplots}
\usepackage{subcaption}
\graphicspath{{./fig/}}
\usetikzlibrary{arrows, calc,intersections, shapes}
\usetikzlibrary{arrows.meta}
\usetikzlibrary{decorations.pathmorphing}
\usetikzlibrary{decorations.text}
\usetikzlibrary{decorations.pathreplacing}
\usetikzlibrary{hobby,decorations.markings}
\usetikzlibrary{positioning}
\usetikzlibrary{calc}
\tikzset{
	every neuron/.style={
		circle,
		draw,
		minimum size=0.15cm
	},
	neuron missing/.style={
		draw=none, 
		scale=1,
		text height=1,
		execute at begin node=\color{black}$\vdots$
	},
}
\input{math_commands}

\pgfplotsset{
	compat=newest,
	plot coordinates/math parser=false,
	every axis/.append style={
		xlabel shift=-0.2em,
		ylabel shift=-0.1em
}}

\usepackage{booktabs}

\usepackage{tikz}
\usepackage{pgfplots,pgfplotstable}
\usepackage{filecontents}

\usepackage[switch]{lineno}  
\usepackage{url}

\def\BibTeX{{\rm B\kern-.05em{\sc i\kern-.025em b}\kern-.08em
		T\kern-.1667em\lower.7ex\hbox{E}\kern-.125emX}}
	
\usepackage{pgfplots,pgfplotstable}
\usepgfplotslibrary{groupplots}
\pgfplotsset{compat=1.14}

\pgfplotstableread[col sep=space]{data.csv}{\datatable} 

\pgfplotstablegetrowsof{\datatable}
\pgfmathsetmacro{\Nrows}{\pgfplotsretval-1}
\pgfmathsetlengthmacro{\MyAxisW}{15cm} 

\usepackage{tikz}
\usetikzlibrary{fit,matrix}

\usepackage{csvsimple}

\tikzstyle{level 1}=[level distance=3.5cm, sibling distance=3.5cm]
\tikzstyle{level 2}=[level distance=3.5cm, sibling distance=2cm]

\usepackage{algorithm}
\usepackage{algorithmic}
\usepackage{glossaries}
\usepackage{siunitx}
\usepackage{cite}
\usepackage{amsmath,amssymb,amsfonts}
\usepackage{amsmath} 
\usepackage{bm}      
\usepackage{algorithm}
\usepackage{algorithmic}
\usepackage{glossaries}
\usepackage{siunitx}
\usepackage{graphicx}
\usepackage{textcomp}
\usepackage{caption}
\usepackage{booktabs}
\usepackage[font=small, labelfont=bf, labelsep=space]{caption}
\usepackage{xcolor}
\def\BibTeX{{\rm B\kern-.05em{\sc i\kern-.025em b}\kern-.08em
    T\kern-.1667em\lower.7ex\hbox{E}\kern-.125emX}}
\makeglossaries
\loadglsentries{gls}
\usepackage{siunitx}
\usepackage[justification=justified,singlelinecheck=false]{caption}
\usepackage[labelsep=colon]{caption}

\begin{document}
\title{\LARGE \bf Uncertainty-Aware Hybrid Machine Learning in Virtual Sensors\\ for Vehicle Sideslip Angle Estimation}

\author{
	 Abinav Kalyanasundaram$^{1}$, Karthikeyan Chandra Sekaran$^{1}$, Philipp Stäuber$^{2}$, \\Michael Lange$^{2}$,  Wolfgang Utschick$^{3}$ and Michael Botsch$^{1}$
	\thanks{$^{1}$CARISSMA Institute of Automated Driving, Technische Hochschule Ingolstadt, Germany, {\tt\small firstname.lastname@thi.de}}%
	\thanks{$^{2}$GeneSys Elektronik GmbH, Offenburg, Germany, {\tt\small lastname@genesys-offenburg.de}}%
    \thanks{$^{3}$Technische Universität München, Germany, {\tt\small utschick@tum.de}}%
}

\maketitle

\def \vneg {-0.45cm}

\begin{abstract}
Precise vehicle state estimation is crucial for safe and reliable autonomous driving. The number of measurable states and their precision offered by the onboard vehicle sensor system are often constrained by cost. For instance, measuring critical quantities such as the \gls{vsa} poses significant commercial challenges using current optical sensors. This paper addresses these limitations by focusing on the development of high-performance virtual sensors to enhance vehicle state estimation for active safety. The proposed \gls{uahi} architecture integrates a machine learning model with vehicle motion models to estimate \gls{vsa} directly from onboard sensor data. A key aspect of the \gls{uahi} architecture is its focus on uncertainty quantification for individual model estimates and hybrid fusion. These mechanisms enable the dynamic weighting of uncertainty-aware predictions from machine learning and vehicle motion models to produce accurate and reliable hybrid \gls{vsa} estimates. 
This work also presents a novel dataset named \gls{revsted}, comprising synchronized measurements from advanced vehicle dynamic sensors.
The experimental results demonstrate the superior performance of the proposed method for \gls{vsa} estimation, highlighting \gls{uahi} as a promising architecture for advancing virtual sensors and enhancing active safety in autonomous vehicles.
\end{abstract}

\section{Introduction}
Safe autonomous driving and \gls{adas} require an accurate estimation for the ego vehicle state. The knowledge about the dynamic states of a vehicle acts as a priori for downstream tasks such as autonomous emergency braking, electronic stability control, and path planning. Among these dynamic states, the \gls{vsa} denoted as $\beta$, is a key indicator of vehicle stability and rider comfort making it crucial for control strategies in active safety systems~\cite{dynamicneuralnetworks}. However, sensors that directly measure~\gls{vsa} for real-time monitoring are expensive to implement in production cars~\cite{Ziaukas}. Alternatively, virtual sensors can estimate vehicle dynamic states by leveraging data from available sensors without space and cost constraints. The \gls{vsa} is often estimated through model-based virtual sensors employing different versions of the \mbox{\gls{kf}}. The recent trend towards software-defined vehicles has paved way for the adoption of \gls{ml}-based virtual sensors~\cite{nox_sensor}. This work presents a virtual sensor designed for reliable~\gls{vsa} estimation from onboard sensor data.

\gls{sota} methods in virtual sensors for \gls{vsa} estimation use observer-based algorithms~\cite{modelbasedvsasurvey,interactingkalmanfilter} or \gls{ml}-based approaches~\cite{dynamicneuralnetworks,Ziaukas}. Observer-based methods struggle to capture the non-linearity in vehicle dynamics and their performance relies on external correction data~\cite{correctionbasedslip}. While \gls{ml}-based approaches can handle non-linearities, they suffer from the “long-tail” problem, where more erroneous predictions occur in underrepresented scenarios. 
They also typically lack the ability to quantify the uncertainty in their predictions~\cite{uncertaintysurvey}. In safety-critical applications, such as autonomous driving, understanding these uncertainties is essential for ensuring reliable estimates from virtual sensor outputs~\cite{marion}.
\begin{figure}[tb]
    \centering
        \includegraphics[scale=1, trim=20pt 0pt 20pt 0pt, clip]{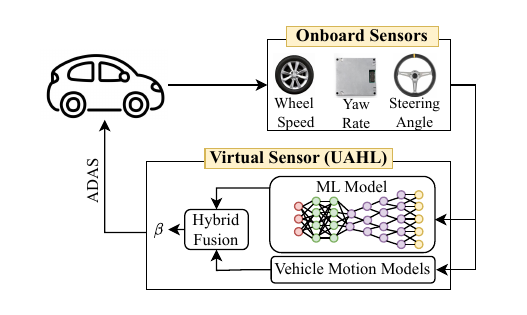}
    \caption{Overview of the proposed \gls{uahi} architecture for virtual \gls{vsa} sensor.}
    \label{fig:uahi_architecture}
    \vspace{-0.5cm}
\end{figure} 
This work aims to address these limitations by introducing the~\gls{uahi} architecture for virtual sensors. It consists of a \gls{ml} model in parallel with vehicle motion models as shown in Fig.~\ref{fig:uahi_architecture}. The \gls{ml} branch considers the \gls{vsa} estimation as a time series forecasting task from onboard sensor data, while the model-based approaches incorporate vehicle dynamics knowledge into the architecture to reduce errors in “long-tail” scenarios. Uncertainty quantification is implemented independently in each branch to leverage their complementary strengths during hybrid fusion. This uncertainty-aware hybrid fusion improves the performance and reliability of estimates from the virtual sensor. 
The main contributions of this work are as follows:
\begin{itemize}
\item Introduction of a novel \gls{uahi} architecture for a reliable virtual \gls{vsa} sensor, with the following components:
\begin{itemize}
    \item[a)] A~\gls{ml} model to estimate \gls{vsa} and its uncertainty based on the Informer~\cite{Informer}.
    \item[b)] \gls{vmms} with uncertainty quantification for its \gls{vsa} estimates.
    \item[c)] Hybrid fusion strategies to combine \gls{ml} and \gls{vmms}.
\end{itemize}
\item Publication of a novel dataset named \gls{revsted} to support research in virtual sensors.  \footnote{Dataset and code to replicate the results will be made open-source upon
acceptance. A dataset sample is provided for reference in \url{https://github.com/MB-Team-THI/UAHL-RevStED}}
\item Evaluation of the performance and reliability of \gls{uahi} architecture as a virtual sensor for \gls{vsa} estimation. 
\end{itemize}

\section{Related works}
Virtual sensors are an active area of research in intelligent vehicles, enabling advancements across a variety of domains, including vehicle dynamics~\cite{activesuspension}, battery thermal management~\cite{batterythermal}, electric drivetrain~\cite{drivetrain}, etc.
They have huge potential in \gls{vsa} estimation, defined as the angle $\beta$ between a vehicle’s longitudinal axis and the direction of its velocity vector at the \gls{cog}~\cite{Ziaukas}. The existing research works on virtual estimation of the \gls{vsa} can be categorized into observer-based approaches and data-driven \gls{ml} methods~\cite{vehiclesideslipsurvey}.

The observer-based approaches like the Luenberger observer, the sliding mode observer and the Kalman filter utilize vehicle models for state estimation~\cite{modelbasedvsasurvey}. They are then combined with measurements from \gls{gps} or inertial measurement units for \gls{vsa} estimation. The vehicle models can be either kinematic~\cite{kinematic1,kinematic2} or dynamic~\cite{vehicledynamics}. While kinematic models offer simplicity from complex parameters, they can suffer in low yaw rate conditions as the system becomes unobservable~\cite{lowexcitation}. Dynamic models incorporate forces and moments acting on the vehicle and vary in complexity depending on the type of model used, such as single-track~\cite{vehicledynamics}, two-track, and tire models~\cite{tiremodel}. Although dynamic models can provide good \gls{vsa} estimation, their accuracy heavily depends on reliable parameters identification, the tire model and the tire wear~\cite{vehicledynamicsproblems}. Recently~\cite{interactingkalmanfilter} attempted to address these challenges using an interacting multiple model \gls{kf} with \gls{gps} data. Many observer-based approaches depend on external correction data~\cite{modelbasedvsasurvey}, which can be challenging to acquire. Moreover, their high accuracy is generally limited to specific vehicle states where the models are most effective~\cite{interactingkalmanfilter}.

The \gls{ml}-based approaches use onboard~\cite{Ziaukas} or external sensor data to directly estimate~\gls{vsa}. Dynamic Neural Networks (DNN)~\cite{dynamicneuralnetworks}, Recurrent Neural Networks~\cite{Ziaukas} and \gls{lstm} models~\cite{lstms} are some of the promising architectures. They exhibited similar performance levels as observer-based approaches in certain maneuvers~\cite{vehiclestateestimationsurvey}. The main challenge with \gls{ml}-based estimation is their inability to quantify uncertainty and performance reliance on dataset quality. In this work, some limitations of both observer-based and \gls{ml}-based approaches are addressed, while leveraging their individual strengths, through the proposed \gls{uahi} architecture for virtual sensors.

\section{methodology}
This section describes the problem formulation, preliminaries and the proposed~\gls{uahi} architecture. The overall architecture is shown in Fig.~\ref{fig:informerarchitecture} and a detailed explanation about each \gls{uahi} component including its integration is provided in the subsequent subsections.
\subsection{Problem Formulation}
Let the dataset $\mathcal{D} = \{(\mX^1,\vy^1),..,(\mX^N,\vy^N)\}$ represent $N$ paired samples. Here, \mbox{$\mX = \{\vx_t\}_{t=1}^L \in \mathbb{R}^{L \times m}$} denotes the history of measured vehicle dynamic states for $L$ observation time steps, where \mbox{$\vx_t = \{x_t^j\}_{j=1}^{m} \in \mathbb{R}^{m}$} denotes the set of measurements obtained from $m$ onboard sensors at time step $t$. The data obtained from a ground truth \gls{vsa} sensor is denoted by \mbox{$\vy = \{y_t\}_{t=L}^{L+F} \in \mathbb{R}^{F+1}$}, where $F$ represents future time steps. The main objective of this work is to develop a function $f_\theta$ to learn the mapping from onboard sensor data $\mX$ towards a high performance \gls{vsa} sensor,
\begin{equation}
\label{equation-1}
    \vy = f_{\theta}(\mX ; L,F,m).
\end{equation}
Here $f_{\theta}$ acts as a virtual sensor by predicting the \gls{vsa} values based on the history of onboard sensor data. 

\subsection{Preliminaries}

The Transformer architecture's~\cite{attention} ability to capture long-range dependencies and interactions, has led to significant progress in time series applications~\cite{transformers_survey}. However, 
the vanilla Transformer~\cite{attention} faced challenges in the time series domain due to the computational complexity of self-attention and its autoregressive decoder~\cite{Informer}. While various strategies have been proposed to modify the network for efficient time series Transformers, the Informer model introduced by Zhou et al.~\cite{Informer} has shown particularly good performance. It explored the low-rank property of the self-attention matrix in time series to introduce \textquotedblleft ProbSparse Self-attention\textquotedblright\;as follows
\begin{equation}
\label{equation-2}
    \mathcal{A}_s(\overline{\mQ}, \mK, \mV) = \text{Softmax}\left(\frac{\overline{\mQ}\mK^\top}{\sqrt{d_k}}\right)\mV,
\end{equation}
here $\overline{\mQ}, \mK, \mV$ represent a sparse query, key and value matrix respectively. The sparse $\overline{\mQ}$ matrix  contains only the top $u$ queries based on sparsity measurement $M(\vq_i,\mK)$ defined as
\begin{equation}
\label{equation-3}
M(\vq_i, \mK) = \max_j \left\{\frac{\vq_i \vk_j^\top}{\sqrt{d}} \right\} - \frac{1}{L_K} \sum_{j=1}^{L_K} \frac{\vq_i \vk_j^\top}{\sqrt{d}} .
\end{equation}
The main intuition behind Eq.~(\ref{equation-3}) is that dominant queries will have corresponding query's attention probability distribution far away from the uniform distribution~\cite{Informer}.
The attention for non-dominant queries is taken as mean over all values~$\mathbf{V}$. This sparse attention mechanism reduces the time and memory complexity of canonical attention from \mbox{$\mathcal{O}(L_Q L_K)$} to \mbox{$\mathcal{O}(\log L_Q L_K)$}, where $L_K$ and $L_Q$ represents the number of rows in matrix $\mK$ and $\mQ$. The proposed \gls{ml} model is based on Informer~\cite{Informer}.
\begin{figure*}[htbp]
\centering
\hspace*{-1.1cm}
\includegraphics[scale=0.91, trim=0pt 0pt 0pt 0pt, clip]{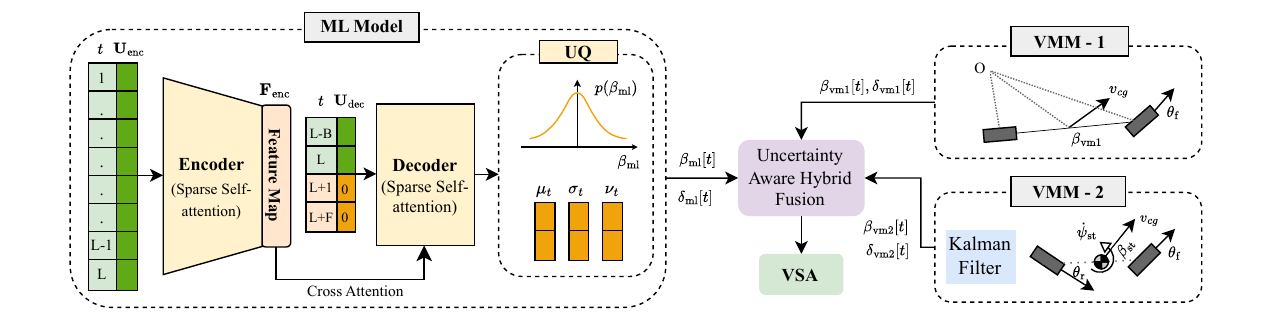}
    \caption{The \gls{uahi} architecture consists of three main components: A \gls{ml} model, \gls{vmms},
and an uncertainty-aware hybrid fusion module. The onboard sensor data $\vx_t$ is the input for both \gls{ml} and \gls{vmms}. The \gls{ml} model, predicts \gls{vsa} $\beta_{\text{ml}}$ with model uncertainty $\delta_{\text{ml}}$ using the Uncertainty Quantification (UQ) block. The \gls{vmms} output their predictions, $\beta_{\text{vm1}}$ and $\beta_{\text{vm2}}$ with their respective uncertainties, $\delta_{\text{vm1}}$ and $\delta_{\text{vm2}}$. Finally, the hybrid fusion module leverages uncertainty-aware inputs to estimate the \gls{vsa} for the virtual sensor.}
\label{fig:informerarchitecture}
\vspace{-0.3cm}
\end{figure*}
\subsection{Machine Learning Model}
\label{sec:ml model}
The proposed \gls{ml} model aims to learn the mapping function $f_{\theta}$ as shown in Eq.~(\ref{equation-1}). Each time step $t$ in the input space $\mX$, represented as $\vx_t$, is transformed to a $d$-dimensional space using an up-projection layer $\Phi$
and added with the positional encoding~\cite{attention} to produce the encoder's input embeddings \mbox{$\mU_{\text{enc}} \in \mathbb{R}^{L \times d}$}. 
The encoder contains two sparse self-attention $\mathcal{A}_s$ blocks with $H$ heads each and a distillation operation~\cite{Informer} between them. 
This distillation operation reduces the temporal dimension $L$ by half. Finally, the encoder outputs a feature map denoted as \mbox{$\mF_{\text{enc}} \in \mathbb{R}^{\frac{L}{2}\times{d}}$} as shown in Fig.~\ref{fig:informerarchitecture}.

The decoder takes as input the last \( B \) observed time steps along with \( F \) future steps, following the approach in~\cite{Informer}. The decoder's input sequence $\mX_{\text{dec}}$ is defined as \mbox{$\{\vx_t\}_{t=L-B}^{L+F} \in \mathbb{R}^{(B+F+1) \times m}$}. Since the \gls{ml} model does not have access to future vehicle states, they are initialized as \mbox{$\{\vx_t\}_{t=L+1}^{L+F} = 0$}. Similar to the encoder, \mbox{$\mX_{\text{dec}}$} is transformed to a \( d \)-dimensional space with the up-projection layer $\Phi$ and added with the positional encoding to form the decoder's input embeddings, \mbox{$\mU_{\text{dec}} \in \mathbb{R}^{(B+F+1) \times d}$}. A sparse self-attention followed by a normal cross-attention~\cite{attention} block is utilized in the decoder to make predictions. The output from the decoder at time steps \mbox{$t \in [L, L+F]$} corresponds to \gls{vsa} estimates as illustrated in Fig.~\ref{fig:informerarchitecture}.

Two major network modifications were made to the Informer~\cite{Informer} to adapt it from time series forecasting to the virtual sensor setting, referred to as Informer*. First, unlike in multivariate time series forecasting, the virtual sensor should not have access to the history of \gls{vsa} values in $\mX$. Therefore, measurements $\vy$ from the ground truth \gls{vsa} sensor were used only during loss computation. Second, uncertainty estimation $\delta_{\text{ml}}$ was incorporated into the model to make reliable \gls{ml} predictions.

\subsubsection*{{Uncertainty Quantification in \gls{ml} (Informer*)}} It was assumed that the \gls{vsa} estimates from the \gls{ml} model follow a Student's t-distribution. Its ability to model heavier tails allows it to handle outliers and variability more effectively than the traditional Gaussian distribution~\cite{bishop2006pattern}. The final fully connected layer in the decoder is designed to predict the mean $\mu_t\in\mathbb{R}^1$, the variance $\sigma_t$, and the degrees of freedom $\nu_t$ of the underlying distribution in order to compute the posterior density for \gls{vsa} $(\beta_\text{ml}[t])$ at time step~$t$. 
To ensure numerical stability during training, these parameters are clipped accordingly to $\sigma_t\in\mathbb{R}^+$ and $\nu_t\in[3, \infty]$.
Correspondingly the loss function for one data sample $(\mX,\vy)$ is formulated as a negative log-likelihood loss
\begin{equation}
\!\mathcal{L}_{\text{ml}}\!=\!-\!\sum_{t=L}^{L+F} \!\ln\!\left(\!\frac{\Gamma\left(\!\frac{\nu_t+ 1}{2}\right)}{\sigma_t \sqrt{\nu_t \pi} \,\Gamma\left(\frac{\nu_t}{2}\right)}\!\left(1\!+\! \frac{(\evy_t\! -\!\mu_t)\!^2}{\nu_t \sigma_t^2}\!\right)^{\!\frac{-\nu_t-1}{2}}\!\right)
\end{equation}
The underlying principle of this loss formulation is that it encourages the \gls{ml} model to output higher variance for more erroneous predictions. 
During testing, the prediction $\beta_\text{ml}[t]$ and uncertainty $\delta_\text{ml}[t]$ for virtual \gls{vsa} estimates are computed as follows 
\begin{equation}
\label{equation:abc}
\beta_{\text{ml}}[t] = \mu_t\;,\;  \delta_{\text{ml}}[t] = \max\left(\frac{\nu_t}{\nu_t - 2} \times \sigma_t^2, 1\right).
\end{equation}

Unlike previous works in \gls{vsa}, the proposed \gls{ml} model is formulated as a time series predictor, utilizing a large observation window $L$ to capture the vehicle's dynamic state within the~\gls{uahi} architecture. It is also designed with a rolling forecasting setting, enabling it to compensate for computational latencies during \gls{ml} model deployment on the vehicle. 
\subsection{\acrfull{vmms}}
\label{sec:vehiclemotionmodels}
The \gls{vmms} are mathematical models that strive to capture the relationship between the vehicle inputs, geometry and dynamic parameters towards its motion while accounting for underlying non-holonomic constraints. The \gls{vmms} act as \textit{a priori} knowledge for virtual sensors to estimate their target quantity from onboard sensor data. In this work, only simple~\gls{vmms} for~\gls{vsa} estimation are utilized as more complex models require several original equipment manufacturer parameters such as tire-parameters, engine torque, etc. 

\subsubsection{\gls{vmm1}} It is a kinematic model designed for flat vehicle movement, assuming negligible tire slip angles. In \gls{vmm1}, the velocity vectors at the “virtual” wheels are aligned with the rolling direction of the tires~\cite{botsch}. This model assumption is valid under the conditions of slow speed cornering, constant steering ratio and absence of rear axle steering. The \gls{vsa} and yaw rate at time step $t$ for \gls{vmm1} are derived from geometric parameters as follows
\begin{equation}
\begin{aligned}
\label{equation-4-5}
\beta_{\mathrm{vm1}}[t] &= \tan^{-1} \left( \frac{\ell_\text{r}}{\ell} \tan(\theta_{\text{A}}[t]) \right), \\
\dot{\psi}_{\mathrm{vm1}}[t] &= \frac{v[t] \tan(\theta_{\text{A}}[t])}{\ell}\;,\; \theta_{\text{A}}[t] = \frac{\theta_{\text{sw}}}{r_\text{s}}.
\end{aligned}
\end{equation}
here $\ell,\ell_\text{r},\theta_{\text{A}},v,\theta_{\text{sw}},r_{\text{s}}$ represent the wheelbase, the distance from the rear axle to the \gls{cog}, the Ackermann steering angle, the velocity, the steering wheel angle, and the steering ratio respectively. This \gls{vmm1} is a simple motion model based on geometric constraints.

\subsubsection{\gls{vmm2}} utilizes the individual wheel speed data obtained from onboard sensors defined in Eq.~\ref{equation-15} to compute the \gls{vsa}. During cornering the speed difference between the inside and outside wheels can be used to give a rough estimate for the lateral velocity $v_{\text{y}}$ at the \gls{cog}\cite{rajamani}. The longitudinal velocity $v_{\text{x}}$ is approximated as the mean speed of all wheels at time $t$. The \gls{vsa} can then be computed as
\begin{equation}
\label{equation-6}
\begin{aligned}
v_{\text{x}} &= \frac{v_{\text{fl}} + v_{\text{fr}} + v_{\text{rl}} + v_{\text{rr}}}{4},\\
v_{\text{y}} &= \frac{v_{\text{fr}} - v_{\text{fl}} + v_{\text{rr}} - v_{\text{rl}}}{2},\;\; \tilde{\beta}_{\text{vm2}}[t]= \tan^{-1}\left( \frac{v_\text{y}}{v_\text{x}} \right) 
\end{aligned}
\end{equation}

This $\tilde{\beta}_{\text{vm2}}[t]$ suffers heavily due to wheel speed noise and the absence of vehicle dynamics during cornering. To improve the rough \gls{vsa} estimates in Eq.~(\ref{equation-6}), a \acrfull{kf}~\cite{kalman} is implemented for \gls{vmm2}, based on the assumption of a Gaussian-Markovian process for vehicle motion. Vehicle dynamic knowledge is incorporated into \gls{vmm2} by using a linear single-track (st) model for system representation~\cite{botsch}. The state $(\vs)$ and measurement vector $(\vz)$ for the \gls{kf} is defined as:
\begin{equation}
\label{equation-7}
\vs[t] = \big[\beta_{\text{st}}, \dot{\psi_{\text{st}}}\big]^\top\;,\;\boldsymbol{z}[t] = \big[\tilde{\beta}_{\text{vm2}}, \dot{\psi}_{\text{obd}}\big]^\top.
\end{equation}
Here, the measurement vector $\vz$ consists of the~\gls{vsa} estimated in Eq.~(\ref{equation-6}) and yaw rate $\dot{\psi}_{obd}$ from onboard sensors. The noise covariance matrices required for the \gls{kf} can be estimated from the dataset. 
The \gls{kf} for \gls{vmm2} is implemented recursively at the frequency of onboard sensor data rate from the vehicle. The final a-posteriori estimates and their covariance $\mP$ from \gls{kf}, serve as the outputs of \gls{vmm2}
\begin{equation}
\label{equation-8}
\vs[t|t] = \big[\beta_{\text{vm2}}, \dot{\psi}_{\text{vm2}}\big]^\top\;,\;\mP[t|t] = \text{diag}\left[\sigma_{\beta}^{\text{vm2}}, \sigma_{\dot{\psi}}^{\text{vm2}}\right]^\top.
\end{equation}

Previous works in \gls{vsa} estimation often relied on complex vehicle dynamic models and filtering algorithms that required external correction data~\cite{vehiclesideslipsurvey}. In contrast, the proposed \gls{uahi} architecture utilizes simple \gls{vmms} to compute the \gls{vsa} from onboard sensor data, without the need for extensive parameter identification. Although simple, the \gls{vmms} can effectively integrate vehicle dynamics knowledge into the \gls{uahi} architecture.

\subsection{Uncertainty Quantification in \gls{vmms}}
\label{sec:uncertainty in vmms}
The uncertainty in a \gls{vmm} arises from the inherent assumptions made to simplify the model. For instance, \gls{vmm1} performs poorly in \gls{vsa} estimation during high-speed cornering, where tire slip angles cannot be neglected. However, numerically quantifying this uncertainty for the \gls{vmms} during inference is challenging due to the absence of a ground truth. Isermann et al.~\cite{Isermann} proposed a fault diagnosis procedure for onboard sensors based on residuals and fuzzy logic. This foundation is extended by proposing a hypothesis for uncertainty estimation in \gls{vmms}.

Let \mbox{$\vx = \{{x}^1,..,{x}^p\}$} denote the vehicle's ground truth state, and \mbox{$\hat{\vx} = \{\hat{x}^1,..,\hat{x}^p\}$} denote the estimates from a \gls{vmm} for $p$ different state quantities. Then the residual vector \mbox{${\ve}=\{{\eve}^1,..,{\eve}^p\}$} is defined as the error in \gls{vmm} estimates such that \mbox{${\eve}^i = \left|\hat{\vx}^i - {\vx}^i\right|$}.
\newtheorem{hypothesis}{Hypothesis}
\begin{hypothesis}
\label{hyp:epistemic_uncertainty}
The uncertainty $\delta_{\text{vm}}$ of a \gls{vmm} arising from model simplifications propagates to the residual vector~$\ve$. This makes the individual residuals correlated to each other such that \mbox{$\eve^i \not\perp \eve^j \;\forall i,j \in p$}.
\end{hypothesis}
The Hypothesis 1 is empirically validated in section~\ref{section:Evaluation and results} and Fig.~\ref{fig:residual_correlation} by showing that $\eve^{\beta} \propto \eve^{\dot{\psi}}$. In this work, Hypothesis~1 is used to quantify the uncertainty in \gls{vmm1} predictions for the \gls{vsa} by leveraging the residual yaw rate 
\begin{equation}
\label{equation-10}
\delta_{\text{vm1}}(\beta_{\text{vm1}}[t]) \approx \left( {\dot{\psi}_{\text{vm1}}[t] - \dot{\psi}_{\text{obd}}[t]} \right). 
\end{equation}
For \gls{vmm2}, innovation residual from \gls{kf} in Eq.~(\ref{equation-7}) can directly used as an uncertainty such that, \mbox{$\delta_{\text{vm2}}[t]=\sigma_{{\beta}}^{\text{vm2}}$}. 

The uncertainty quantification for \gls{vmms} within the \gls{uahi} architecture incurs minimal computational overhead, making it well-suited for real-time monitoring of virtual sensors. While the \gls{vmms} are relatively simple, their ability to produce uncertainty-aware estimates significantly enhances their effectiveness.

\subsection{Uncertainty-Aware Hybrid Fusion}
In safety-critical applications like autonomous driving, a false sense of certainty for virtual sensor estimates can become hazardous. Hence, identifying model limitations through uncertainty evaluation is one of the key aspects of this work.
The complementary strengths of data-driven and model-based methods developed for virtual sensors can be effectively integrated through hybrid fusion as shown in Fig.~\ref{fig:informerarchitecture}. 
The input to the fusion architecture consists of \mbox{$\vh[t] = [\beta_{\text{ml}}, \delta_{\text{ml}}, \beta_{\text{vm1}} ,\delta_{\text{vm1}} ,\beta_{\text{vm2}} ,\delta_{\text{vm2}}]$} at time step $t$ defined in \mbox{Eqs. (\ref{equation:abc}), 
 (\ref{equation-4-5}), (\ref{equation-7}), and (\ref{equation-10})}. Three different \gls{uahi} fusion strategies are proposed varying on the level of interpretability and complexity. 

\subsubsection{Expert Fusion (EF)} The \gls{ml} (Informer*) model generally performs well; however, its performance decreases in highly uncertain regions. This decline in performance is mitigated by incorporating the estimates from the \gls{vmm}s. The vehicle dynamics domain knowledge is leveraged to select the most suitable \gls{vmm} for fusion. At lower speeds, \gls{vmm1} performs well due to the validity of its model assumptions within this region. Conversely, at higher speeds, \gls{vmm2} exhibits superior performance. The proposed expert fusion method is described mathematically by
\begin{equation}
\label{equation-11}
\beta_{\text{EF}}=\left\{
\!\begin{array}{ll}
\beta_{\text{vm1}}\delta_{\text{ml}}\! +\! \beta_{\text{ml}}(1-\delta_{\text{ml}})\!&\! \delta_{\text{ml}}\! >\! \delta_{\text{th}}\! \land \!v_\text{s} \!\leq\! v_\text{th}, \\

\beta_{\text{vm2}}\delta_{\text{ml}}\! +\! \beta_{\text{ml}}(1-\delta_{\text{ml}})\! & \!\delta_{\text{ml}}\! >\! \delta_{\text{th}}\! \land \!v_\text{s}\! > \!v_\text{th},\\
\beta_{\text{ml}} & \delta_{\text{ml}} \leq \delta_{\text{th}},
\end{array}
\right.
\end{equation}
where $\delta_\text{th}$ denotes the uncertainty threshold for the \gls{ml} model and $v_\text{th}$ denotes the velocity threshold for choosing between the \gls{vmm}s.

\subsubsection{Deep Fusion (DF)} Both \gls{ml} and \gls{vmms} produce uncertainty estimates that differ in scale and characteristics. For example, the \gls{ml} model provides probabilistic estimates that may be prone to overconfidence or underconfidence in certain scenarios. In contrast, the \gls{vmms} generate uncertainty estimates rooted in physical principles, making them more interpretable but less adaptable to complex patterns. To account for these differences, a non-linear mapping is learned through a dedicated module, referred to as the deep fusion network ($f_{\text{DF}}$). This hybrid architecture captures complex relationships and interactions between the outputs of the \gls{ml} and \gls{vmms} to generate an estimate for \gls{vsa}
\begin{equation}
\label{equation-12}
\beta_{\text{DF}} = f_{\text{DF}}(\vh).
\end{equation}
After training the \gls{ml} model, a fusion dataset $\mathcal{D}_f$ is created with  $\vh$ as inputs and $y$ as targets. Each time step forms a data sample in this dataset and the predictions are made for each time step independently. However, the dataset splits are not generated by sampling individual time instances but by sampling entire scenarios. This prevents overly optimistic results that could arise from the high correlation between consecutive time steps in the time series. An L2 loss function is utilized to train the model
\begin{equation}
\label{equation-12a}
\mathcal{L}_{\text{DF}} = \|f_{\text{DF}}(\vh) - y\|_2^2.
\end{equation}

\subsubsection{Gaussian Regression Fusion (GF)} A \gls{gmm} is used to learn the joint distribution $p(\vh,y)$ between the fusion input and \gls{vsa}. Expectation-maximization algorithm is used to determine the parameters of the \gls{gmm}. 
During inference, a \gls{gmr}\cite{GMR} is performed by predicting the conditional distribution \mbox{$p(y\mid \vh)$} using the learnt joint distribution \mbox{$p(\vh, y)$}. The estimate with the maximum \mbox{a-posteriori} probability is assigned as
\begin{equation}
\label{equation-13}
\beta_\text{GF} = \argmax_y (p(y \mid \vh)).
\end{equation}

\section{Dataset}
The proposed \gls{uahi} framework for virtual sensors requires a dataset for training and evaluation. Most existing works on \gls{vsa} estimation rely on data collected through onboard sensors in Hardware-in-the-Loop (HIL) simulations~\cite{lstms} or from external retrofitted sensors~\cite{TUdelft}. However, these datasets are often proprietary, limiting their accessibility. To address this limitation, a publicly available, large-scale dataset is essential for advancing virtual sensors.

This work introduces \gls{revsted}, a 5-hour driving dataset comprising space-time synchronized measurements from a test vehicle (a Smart Fortwo from the CARISSMA outdoor test track~\cite{carissma}), equipped with several vehicle dynamic measurement sensors. The primary sensor, the GeneSys ADMA-G-PRO+ GNSS/Inertial System provides precise accelerometer and gyroscope measurements along with accurate vehicle dynamic state estimates at 100 Hz with 1 ms latency \cite{adma}. Complementing this, a Kistler Correvit S-Motion sensor enables slip-free measurement of longitudinal and transverse speeds, as well as \gls{vsa}, making it ideal for driving dynamics tests. This sensor operates at 500 Hz, with an accuracy of $\pm 0.1^\circ$ and a latency of 6~ms~\cite{kistler_correvit}. Furthermore, the dataset includes onboard sensor data from the test vehicle via a dedicated \gls{can} connection. The process of data extraction and synchronization is described below.

\subsubsection{Data Extraction} The dataset was collected by driving the test vehicle in the test track\cite{carissma} by five test drivers. The drivers performed a variety of driving maneuvers, including slalom, constant-radius turns, step steer, sine with dwell, and double-lane changes, to capture diverse vehicle dynamics and sensor responses. The majority of the data was recorded under bright weather conditions, while some recordings were taken during rainy conditions to add variability. The test conditions were designed to provide a comprehensive dataset, enabling research to improve \gls{adas} functionalities through virtual sensing for vehicle state estimation.


\subsubsection{Sensor Synchronization} 
The data obtained from various sensors must be synchronized to achieve a high-quality dataset. Both the ADMA and Correvit sensors obtain timestamps using \gls{gps} data. The latency for onboard sensor data is assumed to be negligible, with timestamps provided upon reception through a high-precision PTP service available through ADMA. 
To ensure consistency across all sensor data, the sensor messages are down sampled to match the frame rate of the slowest sensor, which in this case is the onboard sensor data. For this, each onboard sensor data point is matched to the closest preceding ADMA and Correvit measurements, ensuring a synchronized and accurate dataset for further analysis.
The Correvit sensor measures velocity at the \gls{poi}. This velocity vector $\vv$ is transformed to the \gls{cog} through rigid-body transformations to achieve spatial synchronization
\begin{equation}
\label{equation-16}
\vv_{\text{CoG}}  = \vv_{\text{PoI}} + \boldsymbol{\omega} \times \vr\;,\;\beta_{\text{CoG}} = \tan^{-1} \left( \frac{v_{\text{CoG}}^\text{y}}{v_{\text{CoG}}^\text{x}} \right),
\end{equation}
where $\boldsymbol{\omega}$ denotes the angular velocity of the vehicle obtained from ADMA and $\vr$ the lever arm connecting the \gls{cog} and the Correvit sensor. State estimates for the ADMA data at \gls{cog} are precomputed and directly obtained from the sensor. In total, approximately 0.9 million data samples were collected at a rate of \SI{50}{\hertz}.

The methodologies developed in this work are evaluated using the \gls{revsted} dataset. Inputs to the proposed methodology are obtained through onboard sensor data
\begin{equation}
\label{equation-15}
\vx = \begin{bmatrix}
v_\text{s}, \delta_{\text{sw}}, \dot{\psi}, a_{\text{y}}, p_{\text{br}}, v_{\text{fl}}, v_{\text{fr}}, v_{\text{rl}}, v_{\text{rr}}
\end{bmatrix}^\mathrm{T},
\end{equation}
where $v_{\textbf{s}}$ denotes speedometer reading, $\delta_{\text{sw}}$ the steering wheel position, \mbox{$v_{\text{fl}}, v_{\text{fr}}, v_{\text{rl}}, v_{\text{rr}}$} the speeds of each wheel, $p_{\text{br}}$ the brake pressure and $a_{\text{y}}$ the lateral acceleration. The ground truth output for the \gls{uahi} framework consists of \gls{vsa} measurements $\vy$, from the Correvit sensor.
The data is divided based on a scenario level into training, validation, and test sets, with a split ratio of 70:10:20 continuous across the time domain.

\section{Experiments and results}
In this section, the experimental setup, implementation details, evaluations and ablation studies are discussed. The following questions are answered: (1)~How to validate the proposed Hypothesis 1? (2)~What is the performance improvement achieved by the \gls{uahi} architecture for virtual sensors? (3)~Do hybrid machine learning architectures enhance reliability in estimation? (4)~How does the virtual sensor perform during vehicle dynamics testing maneuvers? 

\subsection{Experimental Setup}

\subsubsection{Baselines} Three recent data-driven \gls{vsa} estimation methods - Dynamic neural Networks~\cite{dynamicneuralnetworks}, ANN-SSE~\cite{lstms}, and RANN~\cite{Ziaukas} were selected as baselines. To effectively compare the modified \gls{ml} (Informer*) architecture, a vanilla  Informer~\cite{Informer} model was also used for comparison. The models were implemented based on the descriptions in the corresponding papers, wherever the code was not publicly available. 

\subsubsection{Implementation details} The \gls{ml} model has an embedding dimension of \mbox{$d=512$} and utilizes \mbox{$H=8$} sparse-attention heads. The encoder processes an observation window of \mbox{$L=100$}, while the decoder operates with a context window of \mbox{$B=25$} and a forecast window of \mbox{$F=5$} steps. The \gls{ml} model was trained with a batch size of \mbox{$b_1=128$} and a learning rate of \mbox{$lr_1=10^{-4}$} using the Adam optimizer for a total number of $e_1=20$ epochs. For the expert fusion, $\delta_{\text{th}}$ was set at the 90th percentile of $\delta_{\text{ml}}$ observed in the validation dataset, and $v_\text{th}$ was set to \SI{20}{\kilo\meter\per\hour}. 
The deep fusion network was trained with a batch size \mbox{$b_2=1000$}, layer dimensions \mbox{$(6,20,10,1)$} employing ReLU activation functions and a learning rate of \mbox{$lr_2=10^{-3}$} for \mbox{$e_2=100$} epochs. 
The Gaussian fusion was trained based on the \gls{gmr}~\cite{GMR}. All models were trained on a single NVIDIA Quadro RTX 5000 GPU with 16 GB VRAM. 

\subsubsection{Evaluation metrics} 
Popular metrics such as \gls{mse} and \gls{mae}~\cite{modelbasedvsasurvey} are used to evaluate the performance of the proposed methods. Additionally, \mbox{\gls{me} $= \max_{i \in \{1,.., N\}} |y^i - \hat{y}^i|$} metric is used to evaluate the potential worst-case scenarios for virtual sensor models. Here, $\hat{y}^i$ and $y^i$ represent the \gls{vsa} prediction and ground truth for the $i$-th sample, respectively. Since some baselines and \gls{vmms} cannot forecast future estimates, all evaluations are conducted exclusively for \mbox{$t=L$} to ensure fair comparison. A large dataset, which includes numerous vehicle maneuvers with high lateral accelerations and diverse weather conditions, is expected to capture the main factors influencing \gls{vsa} estimation. Thus, the effectiveness of the proposed uncertainty-aware hybrid architecture is evaluated using the \gls{revsted} dataset, which will be made publicly available.

\subsection{Evaluation and Results}
\label{section:Evaluation and results}
\subsubsection{Dependency in error residuals} 
\begin{figure}[htbp]
    \centering
    \includegraphics[scale=0.9
    ]{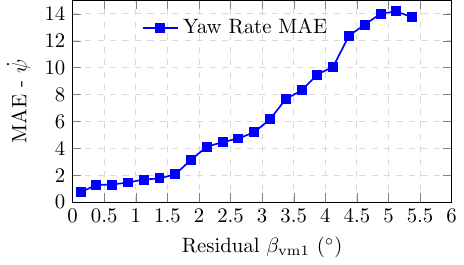}
    \caption{\gls{mae} of yaw rate residual vs \gls{vsa} residual for \gls{vmm1}.}
    \label{fig:residual_correlation}
    \vspace{-0.2cm}
\end{figure}

To validate the proposed Hypothesis 1, the dependency between yaw rate residual $\eve^{\dot{\psi}}$ and \gls{vsa} residual 
 $\eve^{\beta}$ was established for \mbox{\gls{vmm1}}. The Pearson correlation coefficient $r$ quantifying the linear relationship between two random variables and t-test statistic $t^*$ was computed for these residuals. The \gls{vmm1} had \mbox{$r=0.732$} and \mbox{$t^*=453$} for the test dataset. The null hypothesis, which states that no correlation exists between the residuals, was rejected at a 99\% confidence level, providing strong empirical evidence of dependency between the residuals. The residual plot in Fig.~\ref{fig:residual_correlation}, also clearly illustrates that \gls{vmm1} has a linear relationship between yaw rate and \gls{vsa} residual. This empirical validation of Hypothesis 1 supports Eq.~(\ref{equation-10}) and demonstrates that in simple \gls{vmms}, uncertainty estimation for \gls{vsa} can be performed using yaw rate measurements from onboard sensors.
\subsubsection{\gls{uahi} architecture performance for virtual sensors} The proposed \gls{uahi} models outperformed all three baselines, as well as their individual components of \gls{ml} and \gls{vmms} in \gls{vsa} estimation. A quantitative assessment over the test dataset is given in Table~\ref{tab:1} with the best results highlighted in boldface.  
\begin{table}[htbp]
\centering
\vspace{2mm}
\caption{Prediction performance of models in degrees.}
\begin{tabular}{ccccc}
\toprule
\textbf{Models} & \textbf{\gls{mae}} & \textbf{\gls{mse}} & \textbf{\gls{me}} \\ 
\midrule
Dynamic Networks~\cite{dynamicneuralnetworks}      &   0.20    &   0.08   & 3.67   \\ 
ANN-SSE~\cite{lstms}      &  0.21    &  0.10   &    4.09   \\ 
RANN~\cite{Ziaukas}     &   0.19    &  0.09  &   4.19   \\ 
Informer~\cite{Informer}      &   0.18  &  0.06   &   3.04     \\ 
\midrule
\gls{ml}(Informer*)      &   0.16   & 0.12   &  8.23  \\ 
\gls{vmm1}     &    0.41   &  0.58   &    5.32  \\ 
\gls{vmm2}       &   0.25   &  0.15  &  3.36    \\ 
\gls{uahi}-EF    &  0.15    &   0.05   &  2.38    \\ 
\gls{uahi}-DF    &   \textbf{0.12}   &  \textbf{0.03}   &  1.79  \\ 
\gls{uahi}-GF    &  0.14    &  0.04 &   \textbf{1.74}   \\
\bottomrule
\end{tabular}\vspace{-6mm}
\label{tab:1}
\end{table}
Compared to the previous state-of-the-art RANN~\cite{Ziaukas} in~\gls{vsa} estimation, the proposed~\gls{uahi}-DF achieves a 37\% improvement in \gls{mae}. This significant gain can be attributed to both sparse attention in \gls{ml} model and uncertainty-based hybrid fusion. Notably, the 25\% improvement over \mbox{\gls{ml} (Informer*)} highlights the effectiveness of deep hybrid fusion in \gls{uahi}. 

For an in-depth analysis, the vehicle dynamic state $\vx_t$ is conditionally discretized based on speed \mbox{$v_\text{s}(\SI{}{\kilo\meter\per\hour})$} and lateral acceleration \mbox{$a_{\text{y}}(\SI{}{\meter\per\second^2})$}. 
The discretization in $v_\text{s}$ reflects the operation range of \gls{vmm1}, while $a_\text{y}$ captures non-linear vehicle dynamic states. The evaluation results are shown in the Table~\ref{tab:2}.
The \gls{uahi} based on expert fusion performed well in conditions 1 and 3 due to good estimates from the \gls{vmms}. The deep and \gls{gmm}-based hybrid fusion architectures capture the non-linear relationships between different model estimates and their associated uncertainties more effectively. As a result, they produce more accurate fused estimates, even in challenging non-linear vehicle dynamic conditions such as 2 and 4, where the \gls{vmms} fall short. This demonstrates that the \gls{uahi} architecture enhances virtual sensor performance under both normal and challenging vehicle dynamic conditions for passenger cars. 
\begin{table}[h]
\centering
\caption{Conditional prediction performance (\gls{mae} in degrees) of individual \gls{uahi} components and fusion strategies.}
\begin{tabular}{ccccccc}
\toprule
\textbf{VD Conditions} & \textbf{ML} & \textbf{VM1} &\textbf{VM2} &\textbf{EF} & \textbf{DF} & \textbf{GF}\\ 
\midrule
1)$V_s \leq 20 \land a_y \leq 3$      &   0.18  &  0.19 & 0.23 &  0.13    &  \textbf{0.12}  &  0.13  \\ 
2)$V_s \leq 20 \land a_y > 3$     &   0.25  &  1.15 & 0.32 &  0.30    &    \textbf{0.18}  &  0.25 \\ 
3)$V_s > 20 \land a_y \leq 3$      &  0.13 &  0.34 & 0.22  &  0.13   & \textbf{0.13}  &  0.13 \\ 
4)$V_s > 20 \land a_y > 3$     &  0.20  &  1.95 & 0.48 &  0.34   &  \textbf{0.15}  & 0.16 \\  
\bottomrule
\end{tabular}
\label{tab:2}
\end{table}

\subsubsection{Reliability of \gls{uahi} architecture} Since the virtual \gls{vsa} sensor is for active safety applications, ensuring its reliability is crucial. The \gls{me}, quantifying worst-case performance, is a good reliability indicator. All the proposed \gls{uahi} architectures exhibit low \gls{me}, as shown in Table~\ref{tab:1}, indicating their reliability in critical situations. 
Although the \gls{me} for the \gls{ml} model was high, the predicted uncertainty ${\delta_{\text{ML}}}$, was also too high, indicating awareness about its estimate.
To gain deeper insights beyond single-valued metrics, a plot was created to illustrate the distribution of~\gls{mae} across discretized slip angle intervals.
\begin{figure}[htbp]
    \centering
    \vspace{2mm}
    \includegraphics[scale=0.65
    ]{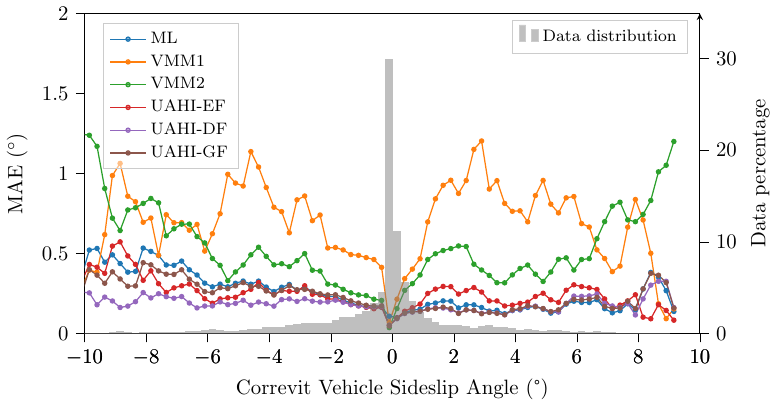}
    \caption{Mean absolute error of \gls{vsa} estimates, calculated for each bin of 0.125° along the ground truth \gls{vsa} $\beta$.}
    \label{fig:correvitslipangleerror}
\vspace{-0.5cm}
\end{figure}
Although \gls{revsted} contains 10\% of data where \mbox{$|a_y| > 3$} due to the deliberate recording of extensive cornering maneuvers, the \gls{vsa} distribution (in \%) still exhibits a “long-tail” pattern, as shown in Fig.~\ref{fig:correvitslipangleerror}. This trend, similar to \cite{TUdelft}, arises because high \gls{vsa} values occur only at the driving limits, which represent only a small fraction of the entire maneuver.
The \gls{uahi} architectures performed significantly better than the individual components for \mbox{$\beta \in [-5^\circ, -10^\circ]$}. At high slip angles \mbox{($\lvert \beta \rvert > 7.5^\circ$)}, which represent corner cases, at least two of the hybrid models outperform the \gls{ml} (Informer*), which suffers from poor estimates due to data imbalance. This shows that \gls{uahi} architecture improves reliability. Additionally, the uncertainty estimation in \gls{uahi} can also support real-time monitoring of the virtual sensor and ensures alignment with ISO 21448~\cite{ISO} standards for risk assessment and the safety of the intended function.

\subsubsection{Performance in vehicle dynamic tests} 
\begin{figure}[htbp]
    \centering
    \includegraphics[scale=0.70
    ]{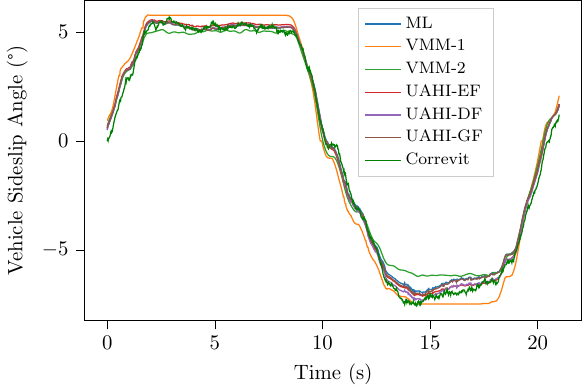}
    \caption{Virtual sensor estimates for \gls{vsa} in ``figure-eight" maneuver.}
    \label{fig: vehicle dynamic tests}
\end{figure}

To evaluate the \gls{uahi} architecture at the limits of vehicle dynamic states, a “figure-eight” maneuver from the test dataset is analyzed and visualized in Fig.~\ref{fig: vehicle dynamic tests}.
It is observed that the \gls{uahi} estimates generally outperform its components, especially in the limiting situation of $\SI{-7}{\degree}$ \gls{vsa}.

\vspace{-0.2cm}
\subsection{Ablation study}
An ablation study was conducted on the best performing \gls{uahi}-DF architecture to assess the contribution of its individual components and the results are shown in Table~\ref{tab:ablation-study}. 
It was observed that fusing the \gls{ml} model estimate with even a single \gls{vmm} estimate yields good performance.
The deep fusion of \gls{vmms} and their uncertainty estimates alone achieved competitive performance compared to the \gls{ml} Informer*. This demonstrates that leveraging the uncertainty patterns of even simple \gls{vmms} can improve \gls{vsa} estimation.
\begin{table}[h!]
\caption{Performance of deep fusion with  selective components.}
\centering
\begin{tabular}{ccccc}
\toprule
\gls{ml} & \gls{vmm1} & \gls{vmm2} & \gls{mae}($^\circ$)  & \gls{me}($^\circ$)\\
\midrule
\checkmark & \checkmark &  &  0.14 & \textbf{1.82} \\
\midrule
\checkmark &  & \checkmark & \textbf{0.13} & 3.35\\
\midrule
 & \checkmark & \checkmark &  0.17 & 3.55\\
\bottomrule
\end{tabular}
\label{tab:ablation-study}
\end{table}
\newpage
\section{Conclusion}
The proposed \gls{uahi} architecture enables the design of a virtual sensor to estimate \gls{vsa} with high precision while relying only on onboard sensor data. Our \gls{ml} model employs a sparse-attention mechanism to effectively capture the vehicle's state over a long history of measurements. Unlike data-driven \gls{ml} methods, \gls{uahi} incorporates vehicle dynamics knowledge through \gls{vmms}, enhancing performance in “long-tail” scenarios. The uncertainty quantification for individual model estimates within the \gls{uahi} architecture enhances its reliability. The introduced hybrid fusion strategies efficiently exploit the uncertainty-aware estimates from the \gls{vmms} and \gls{ml} model to outperform the state-of-the-art approaches in \gls{vsa} estimation. The published \gls{revsted} dataset will also support future works in virtual sensors for vehicle state estimation. 

The \gls{uahi} architecture currently limits the virtual sensor applicability to the test vehicle in the dataset. Future works will focus on understanding the transfer learning to different vehicles. 

\textbf{Acknowledgment:} The authors acknowledge the financial support by the Federal Ministry of Education and Research of Germany (BMBF) in the framework of FH-Impuls (project number 13FH7I13IA).

{
\bibliographystyle{ieeetr}
\bibliography{refs.bib}
}
\end{document}

%% file: math_commands.tex
\def\eqref#1{equation~\ref{#1}}

\def\1{\bm{1}}

\def\ve{{\bm{e}}}

\def\vh{{\bm{h}}}

\def\vk{{\bm{k}}}

\def\vq{{\bm{q}}}
\def\vr{{\bm{r}}}
\def\vs{{\bm{s}}}

\def\vv{{\bm{v}}}

\def\vx{{\bm{x}}}
\def\vy{{\bm{y}}}
\def\vz{{\bm{z}}}

\def\eve{{e}}

\def\evy{{y}}

\def\mF{{\bm{F}}}

\def\mK{{\bm{K}}}

\def\mP{{\bm{P}}}
\def\mQ{{\bm{Q}}}

\def\mU{{\bm{U}}}
\def\mV{{\bm{V}}}

\def\mX{{\bm{X}}}

\DeclareMathAlphabet{\mathsfit}{\encodingdefault}{\sfdefault}{m}{sl}
\SetMathAlphabet{\mathsfit}{bold}{\encodingdefault}{\sfdefault}{bx}{n}

\DeclareMathOperator*{\argmax}{arg\,max}

%% file: gls.tex
\newacronym{adas}{ADAS}{Advanced Driver Assistance Systems}
\newacronym{aeb}{AEB}{Autonomous Emergency Braking}
\newacronym{esc}{ESC}{Electronic Stability Control}
\newacronym{mems}{MEMS}{Micro-Electro-Mechanical System}
\newacronym{vsa}{VSA}{Vehicle Sideslip Angle}
\newacronym{ml}{ML}{Machine Learning}
\newacronym{ai}{AI}{Artificial Intelligence}
\newacronym{vmm1}{VMM-1}{Vehicle Motion Model - 1}
\newacronym{vmm2}{VMM-2}{Vehicle Motion Model - 2}
\newacronym{vmm}{VMM}{Vehicle Motion Model}
\newacronym{vmms}{VMMs}{Vehicle Motion Models}
\newacronym{kf}{KF}{Kalman Filter}
\newacronym{uahi}{UAHL}{Uncertainty-Aware Hybrid Learning}
\newacronym{obd}{OBD}{On Board Diagnostics}
\newacronym{ros}{ROS}{Robot Operating System}
\newacronym{can}{CAN}{Controller Area Network}
\newacronym{pid}{PID}{Parameter IDentification}
\newacronym{gps}{GPS}{Global Positioning Systems}
\newacronym{lstm}{LSTM}{Long Short-Term Memory}
\newacronym{cg}{COG}{Centre of Gravity}
\newacronym{gmm}{GMM}{Gaussian Mixture Model}
\newacronym{revsted}{ReV-StED}{Real-world Vehicle State Estimation Dataset}
\newacronym{sota}{SOTA}{State-of-the-art}
\newacronym{imu}{IMU}{Inertial Measurement Unit}
\newacronym{ins}{INS}{Inertial Navigation System}
\newacronym{dgnss}{DGNSS}{Differential Global Navigation Satellite System}
\newacronym{irs}{IRS}{Inertial Reference System}
\newacronym{adma}{ADMA}{Automotive Dynamic Motion Analyzer}
\newacronym{me}{MaxE}{Max Error}
\newacronym{mse}{MSE}{Mean Squared Error}
\newacronym{mae}{MAE}{Mean Absolute Error}
\newacronym{ptp}{PTP}{Precision Time Protocol}
\newacronym{poi}{PoI}{Point of Installation}
\newacronym{cog}{CoG}{Center of Gravity}
\newacronym{gmr}{GMR}{Gaussian Mixture Regression}